\documentclass[letterpaper, 10 pt, journal, twoside]{IEEEtran}
\usepackage{cite}

\ifCLASSINFOpdf
\else
\fi
\usepackage{url}
\usepackage{graphics} 
\usepackage{epsfig} 
\usepackage{mathptmx} 
\usepackage{times} 
\usepackage{amsmath} 
\usepackage{amssymb}  
\usepackage{subfigure}
\usepackage[T1]{fontenc}
\usepackage[vlined,ruled]{algorithm2e}
\usepackage{booktabs}
\usepackage{multirow}

\usepackage{graphicx}
\usepackage{float}

\usepackage{cite}
\usepackage{xcolor}
\usepackage{algpseudocode}
\usepackage[export]{adjustbox}
\usepackage{balance}
\usepackage{rotating}

\begin{document}
%
\title{Exploiting More Information in Sparse Point Cloud for 3D Single Object Tracking}
%
%
%

\author{Yubo Cui, Jiayao Shan, Zuoxu Gu, Zhiheng Li, Zheng Fang%
\thanks{Manuscript received 25 May 2022; accepted 12 September 2022. Date of publication 22 September 2022; date of current version 30 September 2022}
\thanks{This letter was recommended for publication by Associate Editor G. Costante and Editor E. Marchand upon evaluation of the reviewers’ comments.
This work was supported in part by the National Natural Science Foundation of China under Grants 62073066 and U20A20197, in part by the Science and Technology on Near-Surface Detection Laboratory under Grant 6142414200208, in part by the Fundamental Research Funds for the Central Universities under Grant N2226001, and in part by 111 Project under Grant B16009. (\textit{Corresponding author: Zheng Fang.)}} 
\thanks{Yubo Cui is with the Faculty of Robot Science and Engineering, Northeastern University, Shenyang 110819, China, also with the Science and Technology on Near-Surface Detection Laboratory Wuxi 214000, China, and also with the National Frontiers Science Center for Industrial Intelligence and Systems Optimization, Shenyang 110819, China (e-mail: ybcui21@stumail.neu.edu.cn).}
\thanks{Jiayao Shan, Zuoxu Gu, and Zhiheng Li are with the Faculty of Robot Science and Engineering, Northeastern University, Shenyang 110819, China (e-mail: shanjiayao97@gmail.com; guzuoxu@gmail.com; zhihengli@stumail.neu.edu.cn)}%
\thanks{Zheng Fang is with the Faculty of Robot Science and Engineering, Northeastern University, Shenyang 110819, China, also with the National Frontiers Science Center for Industrial Intelligence and Systems Optimization, Shenyang 110819, China, and also with the The Key Laboratory of Data Analytics and Optimization for Smart Industry (Northeastern University), Ministry of Education Shenyang 110819, China (e-mail: fangzheng@mail.neu.edu.cn).}
\thanks{Digital Object Identifier (DOI): 10.1109/LRA.2022.3208687.}
}
%
%

\markboth{IEEE Robotics and Automation Letters. Preprint Version. Accepted September, 2022}
{Cui \MakeLowercase{\textit{et al.}}: Exploiting More Information in Sparse Point Cloud for 3D Single Object Tracking} 

%



\maketitle

\begin{abstract}
3D single object tracking is a key task in 3D computer vision.
However, the sparsity of point clouds makes it difficult to compute the similarity and locate the object, posing big challenges to the 3D tracker. Previous works tried to solve the problem and improved the tracking performance in some common scenarios, but they usually failed in some extreme sparse scenarios, such as for tracking objects at long distances or partially occluded.
To address the above problems, in this paper, we propose a sparse-to-dense and transformer-based framework for 3D single object tracking.
First, we transform the 3D sparse points into 3D pillars and then compress them into 2D bird's eye view (BEV) features to have a dense representation. 
Then, we propose an attention-based encoder to achieve global similarity computation between template and search branches, which could alleviate the influence of sparsity. 
Meanwhile, the encoder applies the attention on multi-scale features to compensate for the lack of information caused by the sparsity of point cloud and the single scale of features. 
Finally, we use set-prediction to track the object through a two-stage decoder which also utilizes attention. Extensive experiments show that our method achieves very promising results on the KITTI and NuScenes datasets. Code is available at \url{https://github.com/3bobo/smat}.
\end{abstract}


%
\IEEEpeerreviewmaketitle

\section{Introduction}\label{sec:intro}
%
%
%
%
\IEEEPARstart{G}{iven} the initial target object in the first frame, 3D single object tracking aims to estimate the 3D state of the target object in subsequent frames. Therefore, 3D single object tracking has a wide range of applications, such as autonomous driving and robotics. Meanwhile, with the development of 3D computer vision~\cite{PointNet,PointNet++,second,PointPillars,VoxelNet,pv_rcnn}, 3D single object tracking with point clouds received increasing attention. 
Similar to visual tracking, most point-cloud-based 3D single object tracking methods ~\cite{SC3D,P2B,BAT,PTT} also adopt the Siamese pipeline, that is, by cropping the previous and current points based on the previous predicted box to get the template and search point clouds, and then inputting template and search point clouds to predict the state of the object based on their similarity. However, compared to dense images, point clouds are usually sparse, which is not only unfavorable for similarity computation, but also for target localization for 3D single object tracking.
\begin{figure}
	\centering
	\includegraphics[width=\linewidth]{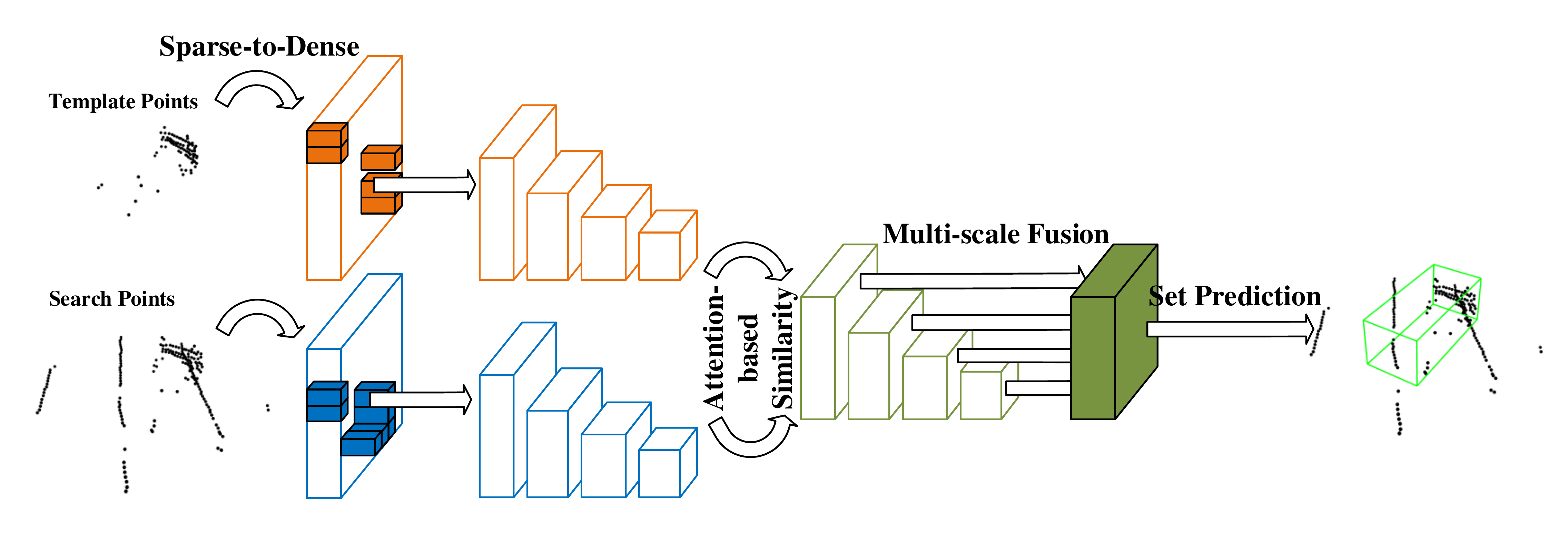}
	\vspace{-0.3in}
	\caption{The overall architecture of our proposed SMAT. We first transform the sparse point cloud to dense BEV features, and use a shared 2D backbone to extract 2D dense features. We then propose an encoder MAE to compute the two branches similarity at each scale and fuse the multi-scale features to fully exploit the information. Finally, we predict the box of object by set-prediction.}
	\label{fig:overall}
	\vspace{-0.2in}
\end{figure}

To address the sparsity problem, previous works usually focus on enhancing the feature representation. 
For example, BAT~\cite{BAT} and V2B~\cite{v2b} enhance the features with box information and shape information respectively, to improve the robustness against sparseness. 
However, since they both adopt a point-based pipeline, they usually need to randomly downsample the points to a fixed number, such as 1024 for the search point cloud and 512 for the template point cloud, to pass through the network, which may lose the geometric information.
LTTR~\cite{lttr} takes voxel-based pipeline and transformer to enhance the features. Nevertheless, they only consider the top-level extracted features but ignore the others, leading to information loss. Meanwhile, their transformer finally outputs region-level features, which are too coarse to address the sparsity problem of point cloud.

Therefore, we believe that a good 3D tracker should have following abilities: \textit{First}, the input representation should retain as much original information as possible. \textit{Second}, because of the sparsity of points, the fusion module should utilize as many features from backbone as possible to have a better similarity. \textit{Third}, the similarity computation should fully exploit the correlation information to have a better measurement.

Based on the above analyses, in this paper, we propose a sparse-to-dense and transformer-based 3D tracking framework, which is named SMAT (\textbf{S}parse-to-dense and \textbf{M}ulti-scale \textbf{A}ttention \textbf{T}racker) and shown in Figure~\ref{fig:overall}.
Specially, we first transform the input points to pillars and apply a 2D backbone to extract features. This transformation could avoid information loss due to downsampling and keep the geometric structure information of the original points. Meanwhile, the extracted feature could have a dense 2D representation, which could alleviate the sparsity problem of the point cloud.
We then propose a multi-scale attention-based encoder to fully exploit the correlation information and compute the similarity. 
Different from previous works that usually ignore the shallow features, we retain all scale features to compute the similarity and then fuse them together. Thus, the similarity could include texture-rich information from shallow features and semantic-rich information from deep features. 
Meanwhile, the encoder uses attention to compute the similarity, which could have more consideration of global correlation information. The multi-head mechanism and global dependence modeling ensure the similarity could have a more comprehensive measurement.
Moreover, for a simple but efficient framework, we adopt an attention-based two-stage decoder to track object by set-prediction.
Comprehensive evaluation results show that our SMAT achieves the state-of-the-art results on KITTI~\cite{KITTI} and NuScenes~\cite{NuScenes} datasets. Overall, our contributions are as follows:
\begin{itemize}	
    \item We propose a sparse-to-dense and transformer-based framework, which adopts an encoder-decoder paradigm, to handle the sparsity challenge in 3D single object tracking.
    
    \item We propose a novel encoder to replace the fusion module in the previous similarity-based tracker~\cite{SC3D,P2B,BAT}, which utilizes the attention at multi-scale features to fully exploit the information from the original sparse point cloud.
    
	\item Our method achieves promising performance on KITTI and NuScenes datasets. Extensive ablation studies also verify the effectiveness of our improvements.
\end{itemize}

The rest of this paper is organized as follows. Section~\ref{sec:related} reviews related work on object tracking and transformer. Section~\ref{sec:methodology} described the overall algorithm framework and details of the network. The experimental results of the proposed method on KITTI and NuScenes datasets are shown in Section \ref{sec:experiments}. Finally, Section \ref{sec:conclusions} concludes the paper.

\section{Related Work}
\label{sec:related}
\subsection{2D Siamese Tracking} 
Recently, the Siamese-like networks~\cite{GOTURN,SiamFC,SiamRPN,ATOM,DiMP,prDiMP,SiamFC++,transt,DualTFR} have been widely used in visual object tracking. The Siamese-like networks usually have two branches for template and search, and extract their features with a shared backbone at first. Then, they fuse the features from two branches by computing their similarity, such as cross-correlation, and use the fused features to regress boxes. However, because of the difference between point clouds and images, these methods are inapplicable to 3D object tracking with point clouds.

\subsection{3D Single Object Tracking}
SC3D~\cite{SC3D} is the pioneering work in 3D single object tracking with point clouds. 
They generate a set of candidate point clouds by Kalman Filter and select the tracked one based on a cosine similarity score.
However, SC3D only takes a one-dimensional feature to compute the similarity, thus losing much local information, especially for sparse point cloud. Meanwhile, the KF also makes it could not be trained end-to-end.
P2B~\cite{P2B} proposes a feature augmentation to augment the point-wise cosine similarity with target cues and takes VoteNet~\cite{votenet} to regress the box. Lately, based on P2B~\cite{P2B}, PTT~\cite{PTT} utilizes the transformer to enhance the fused feature of sparse point cloud, BAT~\cite{BAT} exploits the 3D box information by introducing the box cues into comparison to have an accurate similarity compassion for sparse point cloud. 
However, their random downsampling for the input point cloud may lose the geometric information and makes it hard to output high-quality proposals in sparse scenarios. 
V2B~\cite{v2b} introduces the shape information into features and converts the augmented feature to a dense feature map by voxelization and max-pooling to have dense predictions. 
Nevertheless, they just convert the feature but ignore the input, thus they also suffer from information loss due to the random downsampling.
Additionally, LTTR~\cite{lttr} adopts a voxel-based pipeline to avoid sampling and uses a region-level transformer to enhance the points features. Nonetheless, the region-level enhancement is also too coarse to alleviate the sparsity of the point cloud. Similar as PTT~\cite{PTT}, PTTR~\cite{pttr} also utilizes self-attention to enhance the point features, they further uses cross-attention to compute the similarity between template and search points. Recently, $M^2$-Tracker~\cite{beyond} introduces a motion-based paradigm to track the 3D object, they predicts the relative target motion rather than computing the similarity and achieve the state-of-the-art performance.
	
\subsection{Visual Transformer}
Transformer~\cite{Transformer} is first proposed in natural language processing. With the help of the attention mechanism, the transformer shows a strong ability in modeling the global dependencies of input. Therefore, transformer has also become popular in many tasks of computer vision. ViT~\cite{VIT} applies a pure transformer architecture in image classification. They split the image into many patches and input them into an encoder to classify. 
Swin~\cite{swin} proposes a shifted window-based attention and a pure hierarchical backbone.
PVT~\cite{pvt} designs a progressive shrinking pyramid and spatial-reduction attention to build a pure transformer backbone. They further update PVT to PVTv2~\cite{pvtv2} by overlapping patch embedding and convolutional feed-forward networks. 
Additionally, LocalViT~\cite{LocalViT} introduces convolution into the transformer architecture to improve the local relation modeling. ConvSteam~\cite{Early} finds that convolutional stem could help optimization stability and improve performance for vision transformer. In the dense prediction, DETR~\cite{DETR} first applies the transformer to object detection and deals with the detection problem as a set of predictions to match object queries based on the attention module. Deformable-DETR~\cite{DeformableDETR} further proposes multi-scale deformable attention module to speed up the convergence of DETR and achieve better performance. Meanwhile, transformer has also been introduced into multi-object tracking~\cite{transtrack,trackformer,motBeyond} and segmentation~\cite{SOLQ,segformer,Segmenter}.

\begin{figure*}[t]
	\centering
	\includegraphics[width=\linewidth]{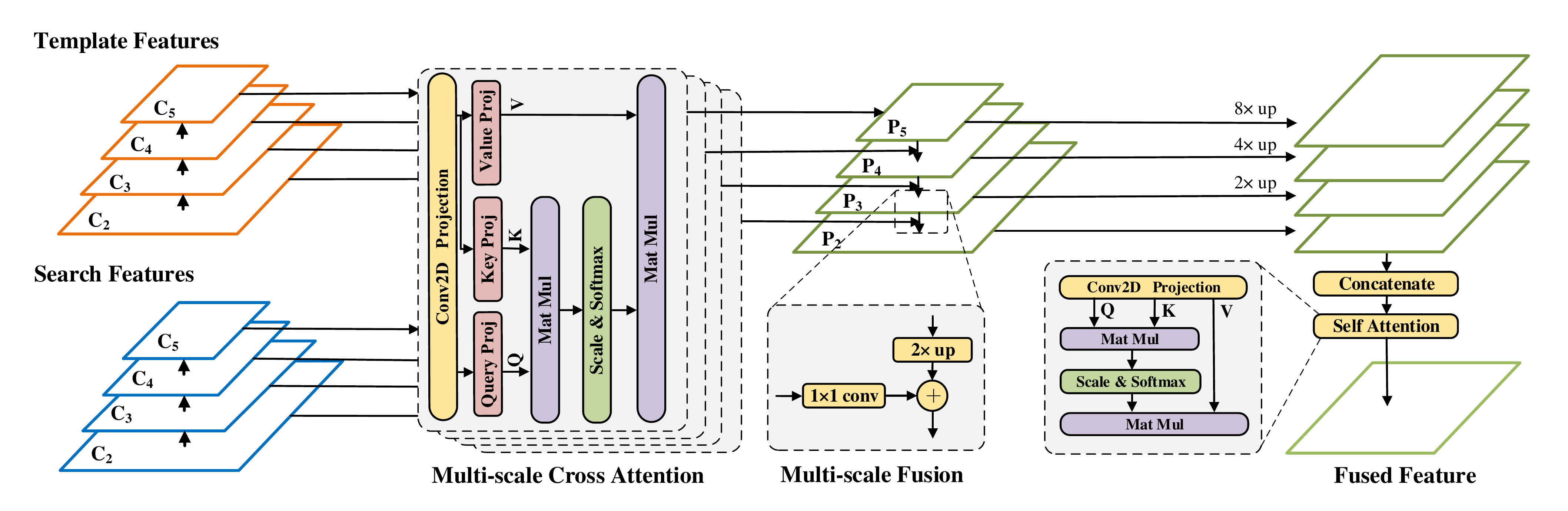}
	\vspace{-0.3in}
	\caption{Illustration of the proposed MAE. Given two sets of multi-scale features from template and search respectively, MAE first computes similarity by attention at each feature scale, and then fuses the multi-scale similarity feature to exploit the information of point cloud.}\label{fig:encoder}
	\vspace{-0.2in}
\end{figure*}

\section{Methodology}
\label{sec:methodology}
\subsection{Overall Architecture}
In the 3D scene, the object box could be represented as $(x,y,z,w,l,h,\theta)$, where $(x,y,z)$ is the center, $(w,l,h)$ is the size and $\theta$ is the orientation of the box respectively. For the tracking problem, our goal is to localize the target object with 3D point cloud frame by frame. Meanwhile, following the assumption~\cite{P2B} that the size of the target object is known through the first frame, we only need to estimate $(x,y,z,\theta)$. 

In this paper, we aim at solving the sparsity problem in the point representation, similarity computation and feature utilization, thus we propose a sparse-to-dense transformer-based framework consisting of three components: sparse-to-dense feature encoding, multi-scale attention-based encoder and two-stage decoder. We will introduce each module in the following subsections.

\subsection{Sparse-to-dense Feature Encoding}
To have a dense representation from sparse points, following~\cite{PointPillars}, we project the points in an area with the size of $W \times L \times H$ from both branches into 3D pillars and apply a simplified PointNet~\cite{PointNet} and maxpooling to generate dense BEV features. 
Following a shared 2D backbone, we extract 2D dense features from the BEV features. Specially, we use an attention-based backbone PVTv2~\cite{pvtv2} to better capture the information of the BEV features. 
By this way, we convert the 3D sparse point clouds into dense 3D pillars and further obtain dense and compact 2D features.

\subsection{Multi-scale Attention-based Encoder}\label{sec:encoder}
After obtaining the two branch features $C^{s}, C^{t}$, where $C^{s}, C^{t}$ represent the search and template feature respectively, we aim at fully exploiting the information of them to better capture the global similarity. Therefore, we propose the \textbf{M}ulti-scale \textbf{A}ttention-based \textbf{E}ncoder (MAE), as shown in Figure~\ref{fig:encoder}. 

\textbf{Similarity Computation. }
Different from previous works using geometric similarity~\cite{P2B,PTT,BAT,v2b} or cross-correlation~\cite{lttr}, we use the multi-head attention mechanism to fuse features from two branches for its better global dependence modeling. The attention is proposed in~\cite{Transformer}, which projects the input features into $Q, K, V$ embeddings to fuse them together based on their similarity. Specially, we first project the feature $C^{s}, C^{t}$ to $E^{s}, E^{t}$ by a shared Conv2D layer. Then, we use an attention-based block to fuse the template feature $E^{t}$ and search feature $E^{s}$. Specially, the attention function is formulated as:
\begin{equation}
{\rm Attention(Q,K,V)} = {\rm Softmax}\left(\frac{QK^{T}}{\sqrt{d}}\right)V{\label{equ:attention}}
\end{equation}
where $Q, K, V$ are the query, key and value embedding respectively and $d$ is the feature dimension of the $K$. Meanwhile, the attention-based block~\cite{Transformer} consists of a multi-head attention (MHA) and a feedforward-network (FFN), thus it could be formulated as:
\begin{align}
{\rm MHA(Q,K,V)} &= {\rm Concat}(H_{1}, ..., H_{h})W^{o}{\label{equ:mha}}\\
{\rm FFN(X)} &={\rm Max}(0,W_{1}X+b_{1})W_{2}+b_{2}\label{equ:ffn}
\end{align}
where $H_{j}$ is computed by Eq.~\ref{equ:attention}, representing attention function for $j$-th head and $W^{o}$ is the head linear projection, $h$ is the head number. $W_{1}, W_{2}$ and $b_{1}, b_{2}$ are weight matrices and basis respectively. $X$ is the output of the ${\rm MHA}$ thus the FFN is after the MHA. Specially, we use a cross-attention block for different inputs to the attention. We project the search feature $E^{s}$ to query embedding $Q$ and project the template feature $E^{t}$ to key embedding $K$ and value embedding $V$ as:
\begin{equation}
	Q = E^{s}W^{q}, K = E^{t}W^{k}, V = E^{t}W^{v}
\end{equation}
where $W^{q}, W^{k}, W^{v}$ are the linear projection of query, key and value respectively. Through the cross-attention block, we obtain the fused similarity feature $P$. 

\textbf{Multi-scale Features Fusion. }
For multi-scale two-branch features fusion, there are two scale fusion strategies: early fusion and late fusion. 
The early fusion first fuses the multi-scale extracted features for each branch and then computes the similarity between two branches and fuses them, while the late fusion first computes two-branch features similarity and fuses them together at each scale and then fuses the multi-scale similarity features.
Compared to the early fusion, the late fusion could exploit multi-scale feature similarity. For example, the top fusion could explore more semantic similarity while the bottom fusion could explore more texture similarity.
Therefore, we adopt the late fusion and will compare the two strategies later.

For each scale features $C^s_i$ and $C^t_i$ from the backbone with downsample rates of $2^i$, where $i \in \{2,3,4,5\}$, we first use the above similarity computation to obtain the fused feature $P_i$ at each scale. Then, to enhance the bottom  features, we use the top-down path and lateral connections following FPN~\cite{fpn} to propagate the information from the top feature to the bottom feature, which could be formulated as:

\begin{small}
\begin{equation}
P_{i-1} = {\rm Conv_{3\times3}}\{{\rm Conv_{1\times1}}(P_{i-1}) + {\rm Upsample}(P_{i})\}, i\in\{3,4,5\}
\end{equation}
\end{small}

Then, to fuse the multi-scale features, we upsample $P_{3}, P_{4}, P_{5}$ to the size of $P_{2}$ and concatenate them together, and apply a Conv2D layer with $1 \times 1$ kernel to fuse the concatenated features to generate scale-fused feature $U$. Finally, a self-attention block is followed to update the fused feature $U$. The process could be formulated as:
\begin{align}
\hat{P_{i}} &={\rm Upsample}(P_{i}), i\in\{3,4,5\}\\
U &= {\rm Conv_{1\times 1}}\{{\rm Concate}(\hat{P_{i}})\}, \forall{i}\\
\bar{U} &= {\rm FFN}({\rm MHA}(U,U,U))
\end{align}

Therefore, given two sets of multi-scale features representing template and search features respectively, the proposed MAE could output a fused feature. Benefiting from the MHA, the fused feature has a global similarity measurement in different feature spaces. Meanwhile, the multi-scale fusion also makes the fused feature have a cross-scale perception, exploiting more information of sparse point cloud.

\subsection{Two-stage Decoder}
Inspired by DETR~\cite{DETR} and Deformable-DETR~\cite{DeformableDETR}, we also adopt a two-stage decoder to predict the box with the fused multi-scale features $\bar{U}$. 
First, we use the fused feature to generates box pairs $\{b_{p}, c_{p}\}$ by two linear layers in regression and classification branches, thus $b_p$ and $c_p$ is the predicted boxes and scores in the first stage respectively. Second, we select top $k$ output pairs based on their scores and project the selected box value $(x, y, z)$ to an embedding feature and concatenate them with their corresponding feature in the feature $\bar{U}$. The concatenated features are further projected by a linear layer to generate a set of target queries $T$, where each query represents the feature embedding of one potential box. The process can be represented as:
\begin{align}
c_{p} &= {\rm Linear_{1}}(\bar{U})\\
b_{p} &= {\rm Linear_{2}}(\bar{U})\\
\hat{b} &= \{b_{i}\|c_{i} \in {\rm TopK}(c_{p})\}\\
\hat{U} &= \{\bar{U}_{i}\|c_{i} \in {\rm TopK}(c_{p})\}\\
T &= {\rm Linear_{3}}({\rm Concate}(\hat{U}, {\rm Proj}(\hat{b})))
\end{align}
where $Proj$ means the concatenation of cosine and sine representation of the proposals. Meanwhile, the target queries could be created and initialized randomly for a one-stage prediction.
Third, the generated target queries $T$ are fed into a cross-attention block together with $\hat{U}$ as:
\begin{equation}
	Q = TW^{q}, K = \hat{U}W^{k}, V = \hat{U}W^{v},
\end{equation}
Through the cross-attention block, the feature $\hat{T}$ is generated for prediction. Following two parallel linear layers to predict the box and score respectively, the decoder outputs a set of $\{b, c\}^{k}_{i=1}$, where $b \in \{x,y,z, sin\theta, cos\theta\}$, and we select the box with the highest score to track.

Different from Deformable-DETR which uses deformable attention and multi-scale features to predict, we only use single scale feature $\bar{U}$ to predict since it already has included multi-scale information in our encoder. Meanwhile, considering the sparsity of 3D points, it would be better to keep more reference points in attention, thus we use the vanilla attention rather than deformable attention. 

\subsection{Training}
Through the proposed SMAT, we get a set of predictions $y=\{b, c\}^{k}_{i=1}$. Different from detection which has many labels in one batch data and considers one object as one label of data, there is only one label for every batch data in single object tracking. Therefore, suffering from the insufficient ground-truth label data, the convergence could be slow in training and the performance may be affected. 

\begin{figure}[t]
	\centering
	\includegraphics[width=0.8\linewidth]{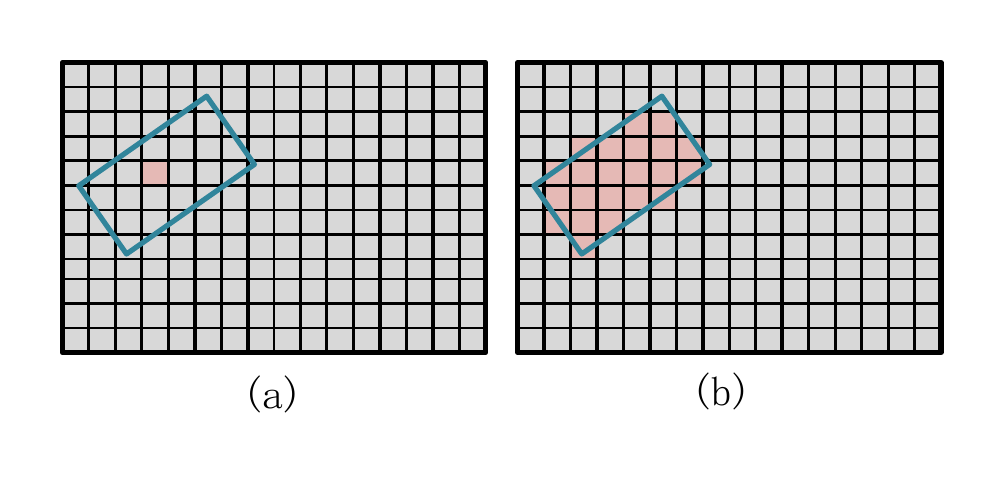}
	\vspace{-0.3in}
	\caption{(a). The previous label assign; (b). Our augmented label assign. Our assign strategy could have more positive samples.}\label{fig:label_assign}
	\vspace{-0.2in}
\end{figure}

To solve this problem, we augment the label data based on the foreground pixels. Specially, we downsample the input point cloud into the same scale with $C_2$ and count the number of foreground pixels $N_{fg}$ in the downsampled BEV point cloud to generate the ground truth set $\tilde{y}=\{\tilde{b},1\}^{N_{fg}}_{i=1}$. The augmentation can be interpreted as every foreground pixel is treated as the object. Figure~\ref{fig:label_assign} shows the difference between previous label assign and our assign. Additionally, we follow the set prediction loss of~\cite{DETR,DeformableDETR} to train the model. The matching cost is defined as follows:
\begin{equation}
	{\cal L} = \lambda_{cls}{\cal L}_{cls} + \lambda_{L1}{\cal L}_{L1}
    \label{eq:matching_loss}
\end{equation}
where ${\cal L}_{cls}$ is cross-entropy loss between the predicted and ground-truth classes, ${\cal L}_{L1}$ is L1 loss between the states of predicted and ground-truth boxes. $\lambda_{cls}$ and $\lambda_{L1}$ are weights for the two losses respectively.

\begin{table*}[t]
\vspace{0.2in}
\renewcommand\tabcolsep{3pt}
\scriptsize
\centering
\begin{tabular}{c|c|cc|cc|cc|cc|cc|cc}
\toprule[.05cm]
\multirow{2}{*}{Method} & \multirow{2}{*}{Paradigm} & \multicolumn{2}{c|}{Car-64159} & \multicolumn{2}{c|}{Pedestrian-33227} & \multicolumn{2}{c|}{Truck-13587} & \multicolumn{2}{c|}{Trailer-3352} & \multicolumn{2}{c|}{Bus-2953} & \multicolumn{2}{c}{Mean-117278} \\
&                           & Success    & Precision   & Success    & Precision   & Success     & Precision    & Success      & Precision    & Success    & Precision & Success    & Precision  \\ \hline \hline
$M^2$-Tracker~\cite{beyond}  & Motion & \textcolor{red}{55.85} &\textcolor{red}{65.09}   &      \textcolor{blue}{32.10}      &  \textcolor{red}{60.92}  &\textcolor{red}{57.36} &\textcolor{red}{59.54}      &  \textcolor{red}{57.61}  &\textcolor{red}{58.26}   &  \textcolor{red}{51.39}  &\textcolor{red}{51.44} &\textcolor{red}{49.23}&\textcolor{red}{62.73} \\ \hline
SC3D~\cite{SC3D} & \multirow{4}{*}{Similarity}    &   22.31         &    21.93  &11.29   &12.65    & 30.67  &27.73    &35.28  & 28.12  & 29.35  &24.08  &20.70&20.20            \\
P2B~\cite{P2B} &   & 38.81  & 43.18  &  28.39 & 52.24  & 42.95 & 41.59&  48.96     &  40.05    &  32.95 & 27.41 &36.48&45.08  \\
BAT~\cite{BAT} &   &40.73 & 43.29 & 28.83 & 53.32   &\textcolor{blue}{45.34}&   42.58   &\textcolor{blue}{52.59}   &\textcolor{blue}{44.89}  &35.44   & 28.01 &38.10&45.71  \\ 

SMAT (Ours) &   & \textcolor{blue}{43.51} & \textcolor{blue}{49.04}  &\textcolor{red}{32.27}  &\textcolor{blue}{60.28} & 44.78& \textcolor{blue}{44.69}&37.45 & 34.10  &\textcolor{blue}{39.42} & \textcolor{blue}{34.32} &\textcolor{blue}{40.20}&\textcolor{blue}{50.92}            \\ \toprule[.05cm]
\end{tabular}
\caption{Performance comparison on the NuScenes dataset. The best two results are highlighted in \textcolor{red}{red}, \textcolor{blue}{blue}}~\label{tab:nuscenes}
\vspace{-0.3in}
\end{table*}

\begin{table*}[t]
\begin{tabular}{c|c|cc|cc|cc|cc|cc}
\toprule[.05cm]
\multirow{2}{*}{Method} & \multirow{2}{*}{Paradigm} & \multicolumn{2}{c|}{Car-6424} & \multicolumn{2}{c|}{Pedestrian-6088} & \multicolumn{2}{c|}{Van-1248} & \multicolumn{2}{c|}{Cyclist-308} & \multicolumn{2}{c}{Mean-14068} \\
           &                           & Success    & Precision   & Success    & Precision   & Success    & Precision   & Success    & Precision   & Success    & Precision   \\ \hline \hline
$M^2$-Tracker~\cite{beyond} & Motion   & 65.5       & 80.8        & \textcolor{red}{61.5}      & \textcolor{red}{88.2}       & \textcolor{red}{53.8}       & \textcolor{red}{70.7}        & \textcolor{red}{73.2} & \textcolor{red}{93.5}  & \textcolor{red}{62.9} & \textcolor{red}{83.4}        \\ \hline
SC3D~\cite{SC3D}      & \multirow{9}{*}{Similarity}                & 41.3       & 57.9        & 18.2       & 37.8        & 40.4       & 47.0        & 41.5       & 70.4        & 31.2       & 48.5        \\
SC3D-RPN~\cite{SC3DRPN}   &                 & 36.3       & 51.0        & 17.9       & 37.8        & - & -  & 43.2       & 81.2        &-  & - \\
P2B~\cite{P2B} &   & 56.2& 72.8 & 28.7 & 49.6 & 40.8  & 48.4  & 32.1 & 44.7 & 42.4       & 60.0 \\
PTT~\cite{PTT}        &                 & 67.8       & \textcolor{blue}{81.8}        & 44.9       & 72.0        & 43.6       & 52.5        & 37.2       & 47.3        & 55.1       & 74.2        \\
BAT~\cite{BAT}        &                 & 60.5       & 77.7        & 42.1       & 70.1        & 52.4       & \textcolor{blue}{67.0}       & 33.7       & 45.4        & 51.2       & 72.8        \\
LTTR~\cite{lttr}      &                 & 65.0       & 77.1        & 33.2       & 56.8        & 35.8       & 45.6        & \textcolor{blue}{66.2}      & 89.9        & 48.7       & 65.8        \\
V2B~\cite{v2b}       &   & \textcolor{blue}{70.5}       & 81.3        & 48.3       & 73.5        & 50.1       & 58.0        & 40.8       & 49.7        & 58.4       & 75.2        \\
PTTR~\cite{pttr}       &                 & 65.2       & 77.4        & 50.9       & \textcolor{blue}{81.6}        & \textcolor{blue}{52.5}       & 61.8        & 65.1       & \textcolor{blue}{90.5}        & 57.9       & 78.1        \\
SMAT (Ours)       &    & \textcolor{red}{71.9} & \textcolor{red}{82.4} & \textcolor{blue}{52.1} & 81.5        & 41.4       & 53.2        & 61.2       & 87.3        & \textcolor{blue}{60.4}       & \textcolor{blue}{79.5}        \\ \toprule[.05cm]
\end{tabular}
\caption{Performance comparison on the KITTI dataset. The best two results are highlighted in \textcolor{red}{red}, \textcolor{blue}{blue}.}\label{tab:kitti}
\vspace{-0.4in}
\end{table*}

\section{Experiments}
\label{sec:experiments}
In this section, we evaluate the proposed SMAT on NuScenes~\cite{NuScenes} and KITTI~\cite{KITTI} datasets. We first introduce the experimental setting and then compare our method with previous state-of-the-art methods on the two datasets. Finally, we conduct extensive ablation studies to investigate each component of SMAT to validate our improvement.

\subsection{Experimental Setting}
\textbf{NuScenes Dataset.}
The NuScenes dataset has a total of 1,000 scenes, contains about 300,000 points every frame and has 360-degree view annotations. Meanwhile, it has been officially divided into training, validation and testing scenes. However, since it does not directly support single object tracking task, we follow the dataset setting of BAT~\cite{BAT} to train and test our method, where we use the training set for training and the validation set for testing. We refer to the published results in~\cite{BAT} for comparison.
	
\textbf{KITTI Dataset.}	
We use the training sequences of KITTI tracking dataset which contains 21 sequences. We follow the same setting as \cite{P2B} to divide the sequences into training, validation, and testing splits, where 0-16 for training, 17-18 for validation and 19-20 for testing.
	
\textbf{Implementation Details.}
We use PVTv2-b2~\cite{pvtv2} as our backbone and follow their original settings.
Specially, the head number is set to [1, 2, 5, 8], the depth is set to [3, 4, 6, 3], the expansion ratio of the feed-forward layer is set to [8, 8, 4, 8] and the channel number is set to [64, 128, 320, 512]. In the Cross-FPN, we set the output channel and feed-forward channel both to 256 and set the cross-attention heads and attention layers to 8. In the decoder, we set the head number and layer number to 8 and the feed-forward channel to 2048. We set $[0.1, 0.1, 4]$m as pillar size and $[-3.2, -3.2, -3, 3.2, 3.2, 1]$m as search area for Car.
	
\textbf{Training Details.}
For KITTI dataset, we train the SMAT with 72 epochs with Adamw~\cite{adamw} optimizer with the initial learning rate of 0.0001, weight decay of 0.05 and batch size of 16 on NVIDIA 3090 GPU. The learning rate decayed by 10× at epochs 63 and 69. For NuScenes dataset, the epochs are 36 with the learning rate decayed by 10× at epochs 27 and 33 and batch size of 32. The other settings are the same as those for KITTI. 
	
\textbf{Evaluation metric.}
We use the One Pass Evaluation (OPE)~\cite{ope} to measure Success and Precision. The Success measures the 3D IoU between the predicted box and the ground-truth box, the Precision measures the AUC of distance between the center of two boxes from 0 to 2m.

\begin{figure}[t]
	\centering
	\vspace{0.1in}
	\includegraphics[width=\linewidth]{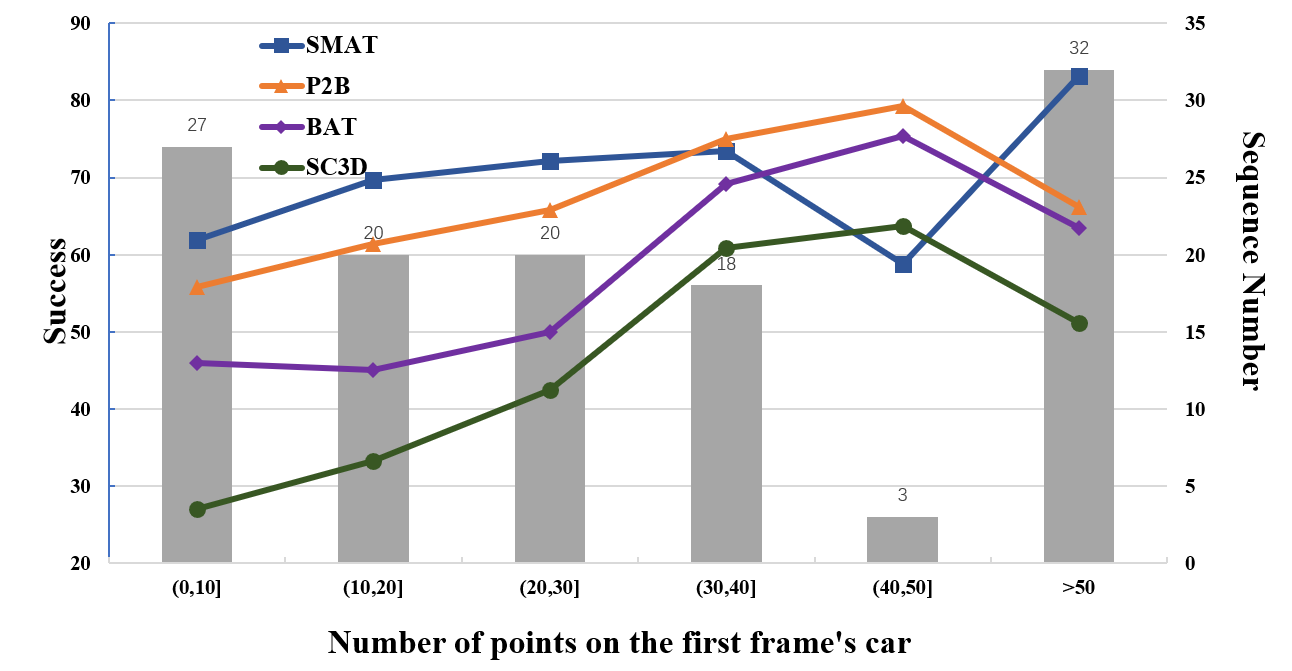}
	\vspace{-0.3in}
	\caption{The influence of the number of points on the first frame’s car.}\label{fig:success}
	\vspace{-0.1in}
\end{figure}

\begin{figure*}[t]
\centering
\vspace{0.1in}
\includegraphics[width=0.9\linewidth]{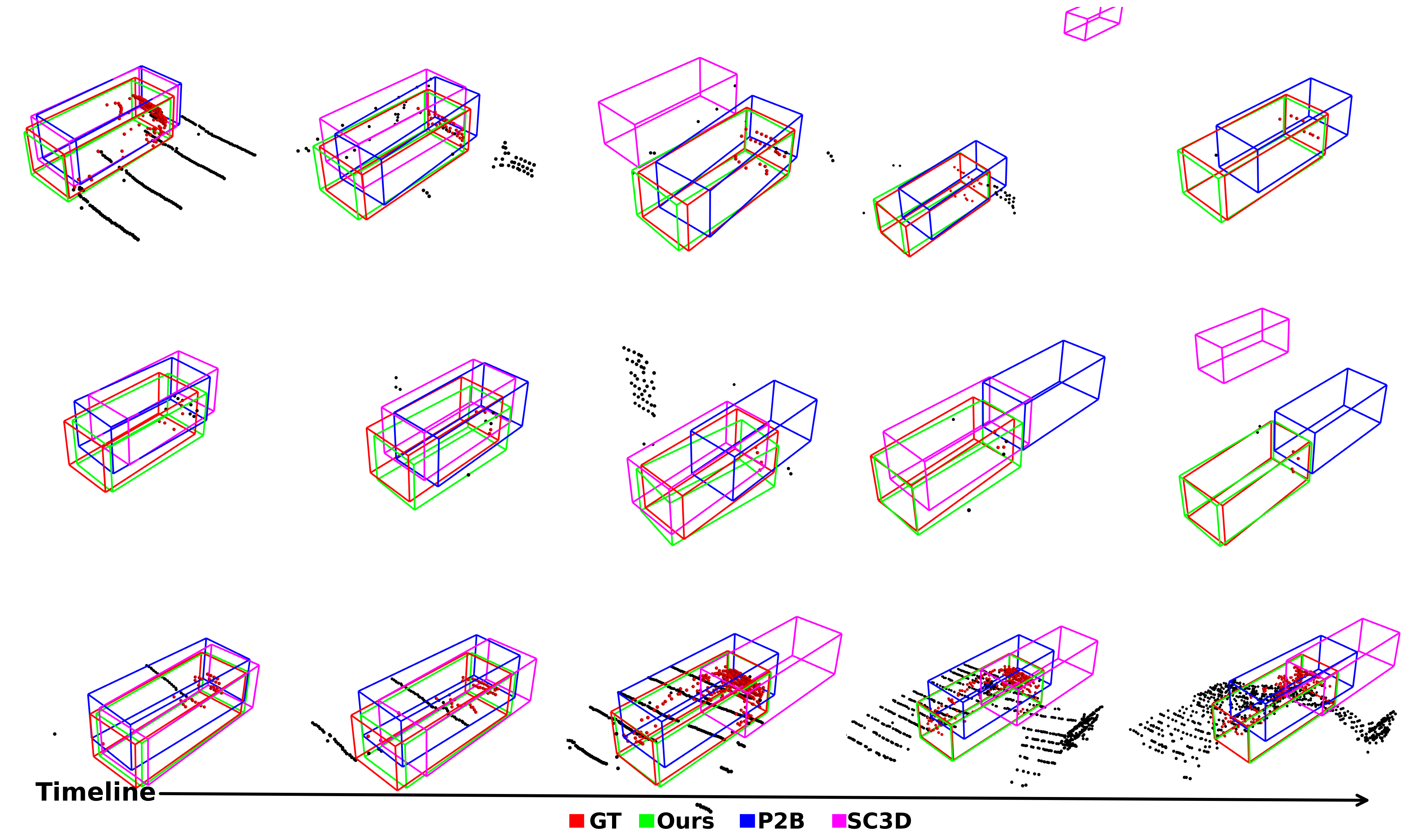}
\vspace{-0.1in}
\caption{Advantageous cases of our SMAT compared with P2B, SC3D on the Car category of KITTI Dataset. The top two rows are sparse while the bottom one is denser. }\label{fig:kitti_results}
\vspace{-0.2in}
\end{figure*}

\subsection{Comparison with State-of-the-arts}
\textbf{Results on NuScenes.} As shown in Table~\ref{tab:nuscenes}, our SMAT achieves the second performance in NuScenes dataset.
Specially, our SMAT lags behind $M^2$-Tracker by 9.03\% in Success and 11.81\% in Precision respectively, and outperforms the third BAT~\cite{BAT} by 2.10\% in Success and 5.21\% in Precision. Compared to our SMAT and other methods, $M^2$-Tracker takes a different motion-based paradigm to track the target, which shows better results than similarity-based paradigm. Meanwhile, our SMAT achieves the best results among the similarity-based methods, showing that our MAE has a better similarity fusion.
Moreover, for the truck class, we have a better prediction on center than BAT~\cite{BAT} but worse prediction on orientation, thus we have achieve better in precision but worse in success.
Additionally, we notice that our trailer’s performance is much lower than other methods. We believe that because of the long and thin shape of the trailer, our method needs a larger search area to input compared to the other methods, thus includes more noise.
\begin{table}[t]
\vspace{0.1in}
	\begin{adjustbox}{center}
		\begin{tabular}{cccccc}\toprule[.05cm]
			Multi-Scale  & Early Fusion & Late Fusion & Success & Precision  \\ \hline \midrule
			$C_{2}$ &      &     & 59.2    & 69.3     \\
			$C_{5}$ &      &     & 61.7    & 74.3     \\
			$\surd$  & $\surd$      &     & 63.8    & 75.7     \\ 
			$\surd$  &       & $\surd$    & \textbf{65.2}    & \textbf{76.2}     \\ \toprule[.05cm]
		\end{tabular}
	\end{adjustbox}
	\caption{Effects of multi-scale feature fusion in the encoder.}\label{table:neck}
	\vspace{-0.35in}
\end{table}

\textbf{Results on KITTI.} As shown in Table~\ref{tab:kitti}, SMAT also performs the second on the mean of four categories. Meanwhile, our method achieves the best performance of 71.9\% and 52.1\% in Car and Pedestrian categories. 
Moreover, SMAT surpasses V2B~\cite{v2b}, which also have a dense prediction, in the Car, Pedestrian and Cyclist categories. The comparison verifies the validity of our sparse-to-dense transformation, showing a dense representation could retain more information from 3D point clouds for single object tracking.
Additionally, compared to LTTR~\cite{lttr} which also adopts a voxel pipeline and transformer, our method exceeds by a large margin (11.7\% in Mean), showing the efficiency of the proposed encoder MAE. Meanwhile, our SMAT achieves 17.6 FPS in the inference phase.

\textbf{Robustness to Sparsity. } To further explore the effectiveness of our method, especially for the sparse point cloud, we classify the car tracking sequences in KITTI according to the number of point clouds in the first frame and then evaluate our method on these sequences of different intervals.
As shown in Figure~\ref{fig:success}, SMAT shows better robustness to sparsity, especially for the targets holding less than 30 points. The figure verifies our improvement in the three aspects. Moreover, only three tracking sequences are in the interval of $(40,50]$, thus we believe that the performance drop in this interval is mainly because of the insufficient samples. Additionally, we also visualize the tracking results in Figure~\ref{fig:kitti_results}.

\subsection{Ablation Study}
In this section, we conduct comprehensive experiments to validate the design of SMAT. All experiments are conducted on the Car category of the KITTI dataset.

\textbf{Multi-scale Fusion in MAE. }
We first analyze the influence of multi-scale feature fusion in our framework. Here we first introduce a baseline network that directly computes single-scale similarity. Meanwhile, the baseline network directly regresses one target box without set-prediction, and we term this regression manner as ``Direct". Notice, for the experiment in Table~\ref{table:neck}, all compared networks use the ``Direct" regression. We believe this experimental setting can more clearly show the effectiveness of our method.

We compared $C_{2}$, $C_{5}$, and multi-scale feature fusion with early and late fusion strategies, where $C_{2}$ and $C_{5}$ are bottom and top features which have downsample rate $2^2$ and $2^5$ respectively. As shown in Table~\ref{table:neck}, compared to using $C_{2}$, using $C_{5}$ performs better by 2.5\% and 5.0\% gains in Success and Precision respectively. Meanwhile, the early multi-scale fusion further improves the performance, surpassing the network with fusing $C_{5}$ by +2.1\% in Success and +1.4\% in Precision. The results show that compared to use single-scale features, fusing multi-scale features could bring better tracking performance in our framework.
Moreover, the late fusion achieves 65.2\% and 76.2\% in Success and Precision respectively, higher than early fusion by 1.4\% and 0.5\%. Compared to early fusion, the multi-scale similarity in late fusion may have a stronger correlation because they only need to focus on the information at their own scale and then fuse the information. In contrast, the early fusion has fused the multi-scale features before the similarity computation, thus the similarity feature only has one scale and may lose some correlation information.

\begin{table}[t]
\centering
\vspace{0.2in}
	\begin{tabular}{c|lcc}\toprule[.05cm]
		\multicolumn{1}{l|}{Prediction} & Similarity        & Success & Precision \\ \hline \midrule
		\multirow{4}{*}{Direct}       & Cosine            & 62.1   & 73.2     \\
		& Euclidean         & 64.6   & 75.0     \\
		& Cross-correlation & 64.1   & 75.6     \\
		& Ours              & \textbf{65.2}   & \textbf{76.2}     \\ \hline  \midrule
		\multirow{4}{*}{Set}            & Cosine            & 66.5   & 76.5     \\
		& Euclidean         & 65.3   & 75.4     \\
		& Cross-correlation & 66.4   & 77.4     \\
		& Ours              & \textbf{71.9}   & \textbf{82.4}
		\\ \hline
		\toprule[.05cm]
	\end{tabular}
	\caption{Comparison of different computations in the encoder.}\label{tab:similarity}
	\vspace{-0.3in}
\end{table}

\begin{table}[t]
\vspace{0.2in}
	\renewcommand\tabcolsep{2pt}
	\begin{adjustbox}{center}
		\begin{tabular}{cccccc}\toprule[.05cm]
			Set-Prediction  & Augmentation & Two-Stage & Success & Precision  \\ \hline \midrule
			&      &     & 65.2   & 76.2     \\
			&      & $\surd$    & 66.4   & 78.4     \\
			$\surd$  &    &  & 53.2    & 65.5  \\
			$\surd$  &       & $\surd$    & 67.0   & 79.1   \\ 
			$\surd$  & $\surd$      &     & 68.8    & 81.4   \\
			$\surd$  & $\surd$      & $\surd$    & \textbf{71.9}    & \textbf{82.4}
			\\ \toprule[.05cm]
		\end{tabular}
	\end{adjustbox}
	\caption{Effects of different components in the decoder.}\label{table:set}
	\vspace{-0.3in}
\end{table}

\textbf{Similarity Computation in MAE.}
We also compare different similarity computation methods in MAE with two prediction manners. We compare the commonly used cosine similarity, Euclidean distance, cross-correlation, and our attention-based computation. Specially, we replace the attention block in MAE with the compared three similarity methods. Meanwhile, the three similarity methods usually need a feature augmentation module in previous works~\cite{P2B,BAT,lttr}, thus we further multiply the three similarity maps with the search feature to serve as a simple augmentation.
As shown in Table~\ref{tab:similarity}, our attention-based method achieves the best performance in both prediction manners, surpassing the second by 0.6\% and 5.4\% in Success of the two prediction manners respectively. The results verify our view that a global similarity computation could have better correlation information for sparse point cloud.
Specially, the geometric similarity only considers one space, such as the feature angle in cosine similarity or the feature magnitude in Euclidean distance. Therefore, the similarity is not sufficient and could be considered as a local measurement in feature space.
Meanwhile, we improve the performance with 1.1\% and 5.5\% in the two prediction manners by replacing the cross-correlation with the attention-based computation.
We believe that the cross-correlation is a spatial local linear matching operation, leading to information loss, especially for the sparse point cloud.
Differently, the attention-based computation explores the global similarity spatially in different feature spaces. Therefore, it fully exploits the correlation information to have a better similarity measurement and bring better tracking performance.

\begin{table}[t]
\vspace{0.2in}
	\centering
	\setlength{\tabcolsep}{4pt}
	\begin{tabular}{l|cccc}
		\toprule[.05cm]
		Method  & Params & FLOPs & Success & Precision  \\ \hline \midrule
		DDETR~\cite{DeformableDETR}&  \textbf{9.4M}    &  31.7G   & 70.9    & 81.4     \\
		Ours  &  10.1M     & \textbf{28.2G}   & \textbf{71.9}    & \textbf{82.4}     \\
		\toprule[.05cm]
	\end{tabular}
	\caption{Comparison with the decoder of Deformable-DETR.}\label{tab:head}
	\vspace{-0.1in}
\end{table}

\begin{table}[t]
\centering
\begin{tabular}{c|c|cccc}\toprule[.05cm]
\multicolumn{1}{l|}{}       & \multirow{2}{*}{Method} & \multicolumn{4}{c}{Source of template}            \\
\multicolumn{1}{l|}{}       &                         & F & P & F\&P & AP \\ \hline
\midrule
\multirow{6}{*}{Success}    & SC3D~\cite{SC3D}   & 31.6  & 25.7  & 34.9 & 41.3\\
                            & P2B~\cite{P2B}    & 46.7  & 53.1  & 56.2  &   51.4       \\
                            & BAT~\cite{BAT}    & 51.8  & 59.2  & 60.5   & 55.8         \\
                            & PTT~\cite{PTT}  & 62.9 & 64.9 & 67.8& 59.8 \\
                            & V2B~\cite{v2b}  & 67.8 & \textbf{70.0} & 70.5 &69.8 \\
                            & SMAT (Ours)  & \textbf{68.1}  & 66.7 & \textbf{71.9} &\textbf{69.9} \\ \hline
\midrule
\multirow{6}{*}{Precision} & SC3D~\cite{SC3D}  & 44.4 & 35.1 & 49.8 &57.9\\
                            & P2B~\cite{P2B}  & 59.7 & 68.9 & 72.8 &66.8\\
                            & BAT~\cite{BAT}   & 65.5 & 75.6  & 77.7 &71.4 \\
                            & PTT~\cite{PTT}   & 76.5 & 77.5  & 81.8 &74.5\\
                            & V2B~\cite{v2b}  & \textbf{79.3} & \textbf{81.3} & 81.3 &\textbf{81.2} \\
                            & SMAT (Ours)  & 77.5  &76.1 & \textbf{82.4} &79.8\\ \hline\toprule[.05cm]
\end{tabular}
\caption{Different template generations. ``F", ``P" and ``AP" denotes the first ground-truth, the previous results and all previous results respectively. The default setting is ``F\&P''. }\label{table:template}
\vspace{-0.3in}
\end{table}

\textbf{Decoder.}
Additionally, we compare the direct regression with set-prediction decoder, analyze the effect of label augmentation and two-stage prediction, and further compare the two-stage decoder with the decoder of Deformable-DETR~\cite{DeformableDETR}. As shown in Table~\ref{table:set}, the performance drops significantly ($\downarrow$12.0\% in Success and $\downarrow$10.7\% in Precision) if we directly adopt one-stage set prediction. 
We believe that because the pure one-stage set prediction only outputs $k$ pair boxes and is trained with only one ground-truth label, thus it is a sparse prediction which is hard to generate high-quality proposals for sparse point cloud.
Meanwhile, the two-stage prediction improves the performance significantly to 67.0\% and 79.1\% in Success and Precision respectively. Additionally, when only using one-stage prediction, the proposed label augmentation also improves the performance to 68.8\% and 81.4\%, showing its simplicity and efficiency.
Finally, by adopting the two-stage set-prediction decoder and label augmentation training together, the proposed network achieves the best performance. Furthermore, the performance of direct regression could be improved to 66.4\% and 78.4\% if we apply two-stage refinement, which is lower than our final result. We believe that direct regression lacks discrimination and forces all features to regress the target value no matter they are from foreground or background, thus limit the performance.
Additionally, as shown in Table~\ref{tab:head}, compared to the decoder in Deformable-DETR~\cite{DeformableDETR}, our decoder achieves better performance ($\uparrow$1.0\% in both Success and Precision). The results show that it is better to keep more reference points rather than reduce them, which is important for 3D object tracking with sparse point cloud.

\textbf{Template Generation Strategy.}
We further compare our method with previous works under different template generation strategies on Car category of KITTI dataset. As shown in Table~\ref{table:template}, our method achieves the best performance in three generation strategies including the first frame ground-truth. The results show that our method could achieve better results when given high confidence prior. Meanwhile, compared to most previous works, such as PTT and BAT, our SMAT has a smaller gap 5.2\% between different strategies, showing more robustness to the source of the template point cloud.

\section{Conclusion}
\label{sec:conclusions}
In this paper, we analyze the limitation of 3D object tracking with point cloud and present SMAT, a sparse-to-dense transformer-based framework. We also propose a multi-scale attention-based encoder MAE to fully exploit the information from the sparse point cloud. By solving the sparsity problem in the point representation, similarity computation and feature utilization, our method could achieve better performance. The comprehensive experiments demonstrate the effectiveness of our design of the proposed framework and encoder. For future work, we plan to explore more attention-based networks on 3D single object tracking.

\ifCLASSOPTIONcaptionsoff
  \newpage
\fi

\bibliographystyle{IEEEtran}
\bibliography{final_version}
\end{document}